\documentclass[11pt]{article}

\usepackage[preprint]{acl}
\usepackage{amsmath}
\usepackage[most]{tcolorbox}
\tcbuselibrary{breakable}
\usepackage{times}
\usepackage{latexsym}
\usepackage{booktabs,multirow}
\usepackage[T1]{fontenc}

\usepackage[utf8]{inputenc}
\usepackage{enumitem}
\usepackage{microtype}
\usepackage{fvextra}
\usepackage{inconsolata}

\usepackage{graphicx}

\title{Empowering Reliable Visual-Centric Instruction Following in MLLMs}

\author{Weilei He$^{1}$ ~~~Feng Ju$^{1}$ ~~~Zhiyuan Fan$^{1}$ ~~~Rui Min$^{1}$ ~~~Minhao Cheng$^{2}$ ~~~Yi R. (May) Fung$^{1}$ \\
  $^{1}$Hong Kong University of Science and Technology  ~~~~~$^{2}$Penn State \\
  \texttt{weileihe090@whu.edu.cn \quad yrfung@ust.hk}}

\begin{document}
\maketitle

\begin{abstract}
Evaluating the instruction-following (IF) capabilities of Multimodal Large Language Models (MLLMs) is essential for rigorously assessing how faithfully model outputs adhere to user-specified intentions. Nevertheless, existing benchmarks for evaluating MLLMs’ instruction-following capability primarily focus on verbal instructions in the \textbf{textual modality}. These limitations hinder a thorough analysis of instruction-following capabilities, as they overlook the implicit constraints embedded in the semantically rich \textbf{visual modality}. To address this gap, we introduce VC-IFEval, a new benchmark accompanied by a systematically constructed dataset that evaluates MLLMs’ instruction-following ability under multimodal settings. Our benchmark systematically incorporates vision-dependent constraints into instruction design, enabling a more rigorous and fine-grained assessment of how well MLLMs align their outputs with both visual input and textual instructions. Furthermore, by fine-tuning MLLMs on our dataset, we achieve substantial gains in visual instruction-following accuracy and adherence. Through extensive evaluation across representative MLLMs, we provide new insights into the strengths and limitations of current models.\footnote{Our code will be released publicly upon publication.}
\end{abstract}
\section{Introduction}
\begin{figure}[t]
    \centering
    \includegraphics[width=\linewidth]{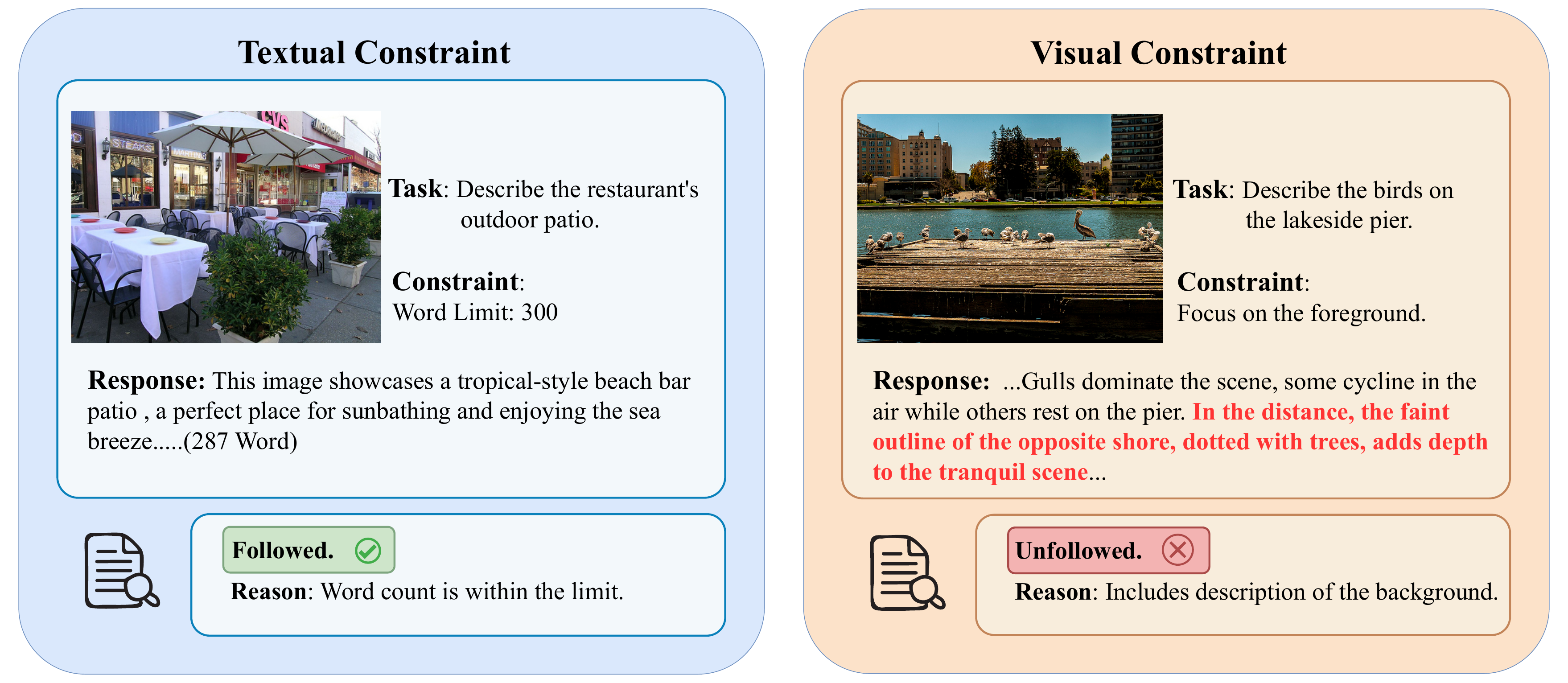}
    \caption{We remove the visual input and prompt the model separately with textual and visual instructions. The response generated under the textual instruction is visually inconsistent with the image, yet it can still pass the instruction-following judgment. In contrast, the response generated under the visual instruction is visually inconsistent with the image and is thus judged by the model as not following the instruction.
    }
    \label{fig:judge}
\end{figure}

Large Language Models (LLMs) have achieved remarkable progress across a wide range of natural language processing tasks, including question answering \cite{wei2022chain}, summarization \cite{ouyang2022training}, dialogue \cite{thoppilan2022lamda}, and code generation \cite{sun-etal-2023-decoding}. In parallel, Multimodal Large Language Models (MLLMs) have extended these capabilities to visual-language settings, enabling applications such as visual question answering, image captioning, and multimodal reasoning \cite{su2025thinking,wu-etal-2024-macaroon}. A fundamental ability that underpins the success of both LLMs and MLLMs is instruction-following (IF)—the capacity to correctly interpret and execute user-provided instructions. Specifically, consider a visual task where the user provides an image of a busy street and asks: \textit{“Describe the scene, but focus only on the actions of pedestrians while ignoring vehicles and buildings.”} A reliable model must not only generate a description consistent with the image but also adhere to the visual constraint in the instruction. For example, the model should produce \textit{“Two people are crossing the street while another is talking on the phone”}, rather than describing irrelevant objects like cars or buildings. This ability is central to ensuring that LLMs and MLLMs are reliable, controllable, and aligned with human intent in real-world applications.

Achieving multimodal instruction-following (IF) remains a highly challenging problem, primarily bottlenecked by the scarcity of high-quality training data and rigorous evaluation benchmarks. Existing multimodal IF datasets and benchmarks \cite{ding2025mm, qian2024mia} suffer from critical shortcomings: they predominantly assess adherence to textual patterns, neglecting the crucial role of visual inputs, as it is shown on Figure.\ref{fig:judge}. Consequently, they often evaluate the IF ability of the LLM backbone rather than the model's capacity for genuine multimodal understanding. Similarly, existing training data \cite{ding2025mm} lacks high-quality, visually grounded examples, and evaluation protocols that rely on text matching fail to adequately measure the contribution of vision. Together, these limitations severely undermine progress in developing robust and reliable MLLMs.

To address the scarcity of high-quality visual instruction-following (IF) data, we propose a novel pipeline to construct visual-centric IF datasets. Our methodology generates and refines visually grounded tasks for images and then creates a broad spectrum of corresponding visual instructions, covering diverse types such as spatial grounding, style specification, and target restriction. Leveraging this pipeline, we build two datasets: \textbf{VC-IFInstruct}, a high-quality dataset of image-response pairs for supervised fine-tuning (SFT), and \textbf{VC-IFDPO}, a preference dataset with contrastive data generated via techniques like instruction ablation for Direct Preference Optimization (DPO) \cite{DPO}.

To facilitate rigorous evaluation of multimodal IF, we introduce VC-IFEval, a benchmark designed to capture the critical role of visual inputs in instruction-following. VC-IFEval contains hundreds of visually grounded questions and dozens of instruction types. Unlike prior benchmarks \cite{ding2025mm, zhou2023instruction} that rely solely on text-based compliance, VC-IFEval employs a novel hybrid evaluation strategy. In the direct evaluation, we use an expert model (GPT-4o) as a judge \cite{lee2024prometheusvisionvisionlanguagemodeljudge} to assess whether responses generated with visual inputs satisfy the given constraints. Furthermore, we adopt a comparative evaluation scheme, where model outputs with and without visual inputs are compared to isolate the contribution of vision, reducing non-visual confounding factors in assessing instruction-following ability. This challenging setting is further evidenced by the fact that even a leading open-source model, Qwen2.5-VL-Instruct-32B \cite{bai2025qwen25vltechnicalreport}, achieves only 67.3\% success rate on VC-IFEval, suggesting substantial room for improvement in multimodal instruction-following.

We further demonstrate the effectiveness of our resources by fine-tuning open-source MLLMs on VC-IFInstruct (SFT) and VC-IFDPO (preference learning). Experimental results show consistent improvements in instruction-following benchmarks \cite{ding2025mm, zhou2023instruction}, validating the effectiveness of our datasets for enhancing IF capability. Furthermore, evaluations on standard visual QA benchmarks indicate that models trained with our data not only achieve stronger instruction adherence, but also maintain or even improve performance on general visual understanding tasks. 

Our contributions are fourfold.
(1) We propose a systematic and fine-grained pipeline for constructing visual instruction data, which introduces diverse types of visual constraints—covering spatial, stylistic, structural, and task-specific aspects—to enhance multimodal grounding.
(2) Based on this pipeline, we build two large-scale datasets: VC-IFInstruct for supervised instruction tuning and VC-IFDPO for preference optimization, together forming a unified training framework for improving visual instruction-following.
(3) We further develop VC-IFEval, a challenging and diagnostic benchmark featuring rich visual instructions and a hybrid evaluation protocol that explicitly isolates the contribution of visual inputs.
(4) Through extensive experiments, we demonstrate that training on our datasets consistently boosts MLLMs’ instruction-following capability while maintaining general visual understanding performance, highlighting the effectiveness and generality of our framework.

\section{Related Works}
\paragraph{Instruction Following in Foundation Models} Instruction following (IF)~\cite{zeng2023evaluating, zhou2023instruction} is a fundamental capability that characterizes the alignment between the outputs of foundation models and user intent. While existing work~\cite{zhou2023instruction, qin2024infobench, jiang2023followbench, zhang2024cfbench, wen2024benchmarking, xia2024fofo} has primarily focused on evaluating instruction following in large language models (LLMs), this capability remains less understood in multimodal contexts. To bridge this gap, several recent studies have shifted toward multimodal large language models (MLLMs)~\cite{hurst2024gpt}, seeking to investigate how the inclusion of visual modalities impacts instruction following performance~\cite{bitton2023visit, qian2024mia, ding2025mm}. While these studies emphasize increasing task diversity~\cite{ding2025mm} and complexity~\cite{qian2024mia}, their benchmarks still primarily assess instruction following within the textual modality, overlooking the implicit cues embedded in the visual modality. To this end, we introduce our VC-IFEval, a challenging benchmark designed for \textbf{visual-centric evaluation} of the instruction-following capabilities of MLLMs. 



\section{VC-IFEngine}

\begin{figure*}[t]
  \centering
  \includegraphics[width=\textwidth]{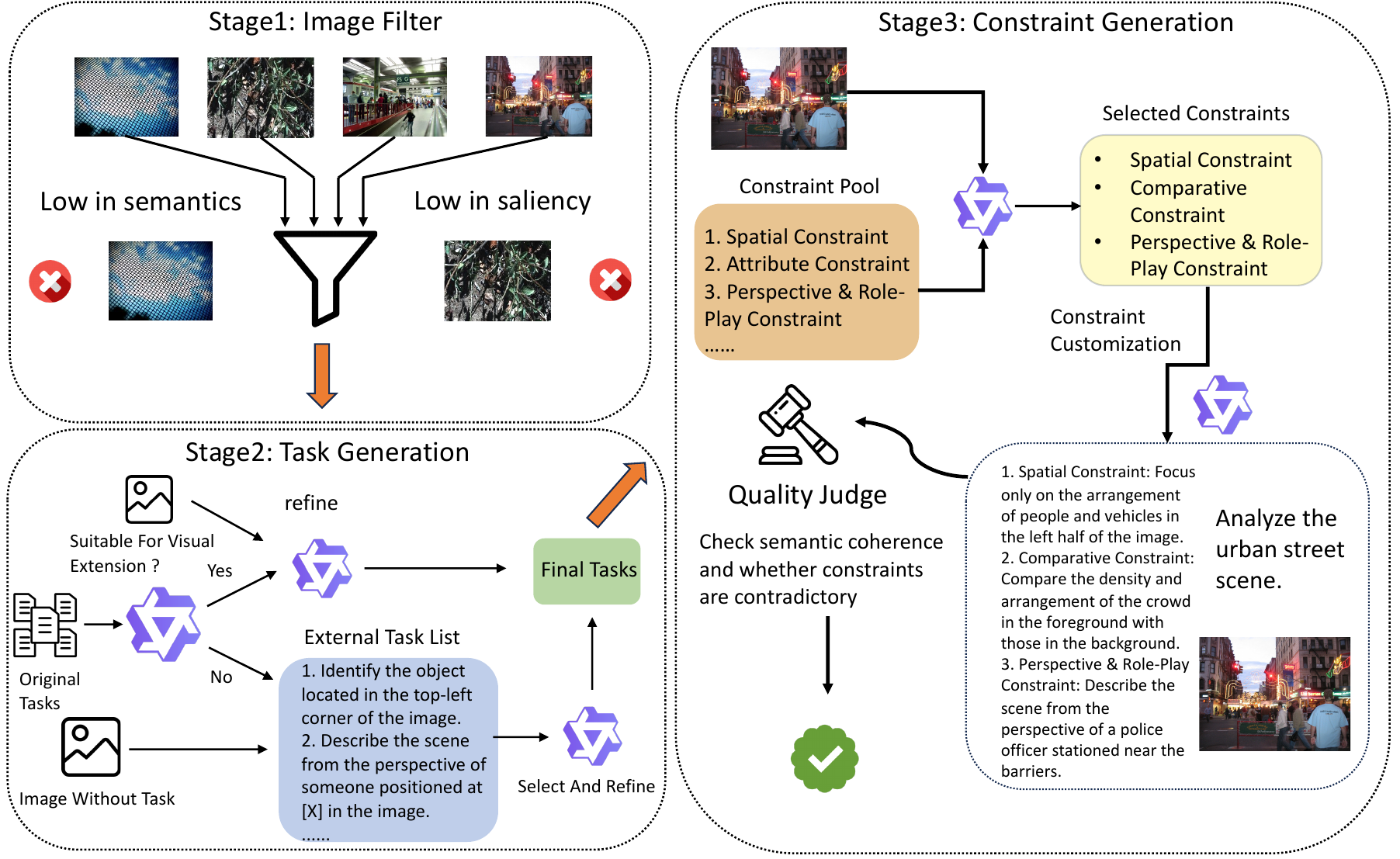}
  \caption{\textbf{Overall pipeline of VC-IFEngine.}
The framework operates in three stages:
\textbf{(1) Image Filter}, which removes visually uninformative samples with low semantics or saliency;
\textbf{(2) Task Generation}, where we refine existing annotations or sample new tasks from an external pool depending on whether images contain predefined tasks;
and \textbf{(3) Constraint Generation}, which selects and customizes constraints from a predefined pool and ensures their coherence through a quality-judging step.
The resulting visually grounded tasks and validated constraints form the foundation for constructing VC-IFInstruct, VC-IFDPO, and VC-IFEval, enabling fine-grained multimodal instruction generation and reliable evaluation.}
  \label{fig:vc-ifengine-pipeline}
\end{figure*}
We propose VC-IFEngine, a novel pipeline designed to construct datasets that emphasize visual-centric instruction-following. As illustrated in Figure~\ref{fig:vc-ifengine-pipeline}, our pipeline ensures high-quality data for both evaluation and instruction tuning through three stages: (1) \textbf{image filtering}, (2) \textbf{task generation}, and (3) \textbf{instruction generation}. Specifically, the pipeline performs image filtering to ensure diverse visual content, task generation to create or refine image-relevant tasks, and instruction generation to produce visual instructions for each image--task pair.
\subsection{Image Filter}
The goal of our image filtering stage is to remove images with low information density and visual saliency, while retaining those that are visually rich and easy to interpret. To quantify the visual information density of each image, we adopt the IC9600 metrics \cite{feng2023ic9600}, which provide an effective measure of content richness. The measurement result is shown in Figure \ref{fig:image_filter_dist}. As shown in the figure, both the mean and median scores are approximately 0.50, indicating a near-symmetric distribution of visual information density across the image pool. Thus, we choose 0.50 as the threshold to filter images. However, filtering solely based on information density is insufficient, as images with overly complex visual content can hinder LVLMs from accurately understanding scenes, refining tasks, and generating reliable instructions.

To address this, we further apply RAM, or the Recognize Anything Model \cite{zhang2023recognize}, to perform object recognition and evaluate the saliency of visual content. By prioritizing images with clearer and more distinguishable visual signals, we ensure that the selected samples are better aligned with the downstream stages of task generation and instruction generation.

\subsection{Task Generation}

To ensure that each image is associated with a set of tasks suitable for visual instruction extension, we adopt a unified task generation framework that handles both images with and without existing task annotations.

For each image $I$, if original task annotations $T_{orig}$ are available, we first examine each task $t \in T_{orig}$ to determine whether it can be effectively extended into a visual instruction. If a task is deemed suitable, it is refined into a clearer, more specific, and visually grounded version $t'$, preserving its original intent while enhancing its alignment with the image content.

If a task is found unsuitable for visual extension, such as being overly abstract, ambiguous, or weakly related to the image, we replace it with a new task drawn from an external task pool $\mathcal{P}_T$. Similarly, for images that lack any original annotations, we also sample candidate tasks from $\mathcal{P}_T$ and refine them according to the visual content of the image.

Formally, given a sampled subset of task candidates $\mathcal{T}_e \subset \mathcal{P}_T$, we prompt a large language model $\mathcal{M}$ to produce a final task list $\mathcal{T}_I$ conditioned on both the image and the task candidates:
\begin{equation}
\mathcal{T}_I = \mathcal{M}(I, \mathcal{T}_e).
\end{equation}

The model $\mathcal{M}$ selects the most relevant tasks or generates refined alternatives that better fit the visual semantics of $I$. More details are shown in Appendix \ref{appendix:task_constraintgeneration}. This unified process guarantees that every image, regardless of whether it has existing annotations, is paired with semantically rich and visually grounded task formulations, establishing a solid foundation for the subsequent instruction generation stage.
\subsection{Constraint Generation}

\paragraph{Constraint Pool}
 To systematically capture the different aspects of visual reasoning, all constraints in our work are organized into two major categories and ten subcategories, each reflecting a distinct perspective of visual understanding. Please refer to Appendix Fig.\ref{tab:constraint_pool} for more detailed examples of these categories.

\paragraph{Constraint Development}
Given the predefined constraint pool and the generated task instructions, we develop image-specific and high-quality visual constraints through a three-stage process: \textbf{selection}, \textbf{refinement}, and \textbf{quality judgment}.  

First, for each image--task pair $(I, T^*)$, we sample a subset of candidate constraints $C_s$ from the global constraint pool, ensuring coverage over various reasoning aspects such as spatial relations, attributes, and comparative logic.  

Next, we prompt a large vision-language model (LVLM) to refine these selected constraint types into concrete, image-grounded formulations:
\begin{equation}
C_r = L(I, C_s, T^*),
\end{equation}
where $L$ denotes the LVLM-based refinement function that incorporates visual and semantic cues to produce detailed and context-aware constraint descriptions. This refinement step adapts the abstract constraint types to the specific image and task, ensuring visual grounding and appropriate granularity.  

Finally, to guarantee the coherence and reliability of the refined constraints, we perform a \textbf{quality judgment} stage using VLM-as-a-judge \cite{lee2024prometheusvisionvisionlanguagemodeljudge}, in which the LVLM re-evaluates each candidate constraint for relevance and consistency:
\begin{equation}
C_f = V(I, C_r, T^*),
\end{equation}
\noindent where $V$ represents the validation function that filters out contradictory, redundant, or irrelevant constraints. For more detailed information, please refer to Appendix \ref{appendix:task_constraintgeneration}
\subsection{Tuning Data Preparation}
\begin{figure}[t]
    \centering
    \includegraphics[width=\linewidth]{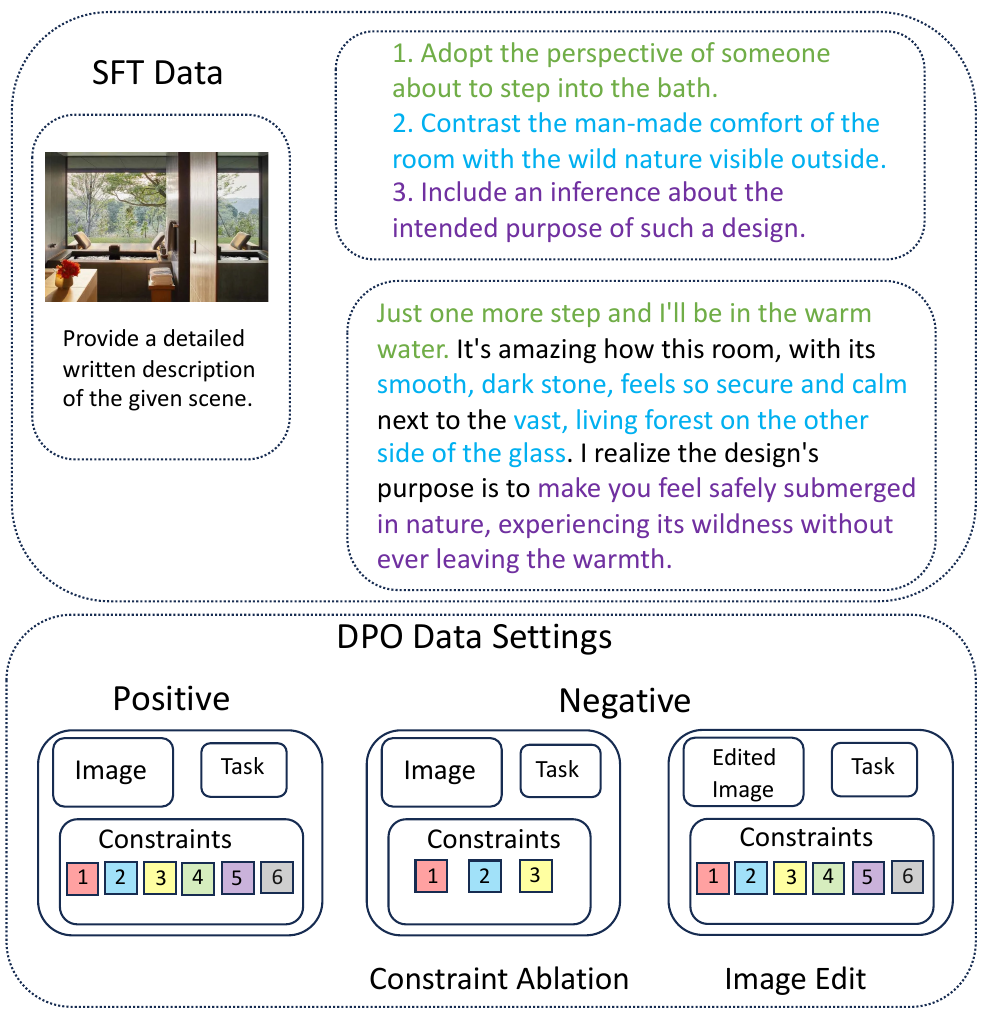}
    \caption{Illustration of SFT and DPO data settings
    }
    \label{fig:data}
\end{figure}
\paragraph{VC-IFInstruct}
By applying the VC-IFEngine pipeline, we construct the VC-IFInstruct dataset, shown in Figure\ref{fig:data}, which provides high-quality visual instruction-following training data for both supervised fine-tuning and preference optimization. To generate model responses, we adopt the Qwen2.5-VL-32B-Instruct model \cite{bai2025qwen25vltechnicalreport}, which demonstrates strong performance on a range of multimodal benchmarks and exhibits competitive instruction-following ability compared to state-of-the-art LVLMs. Finally, VC-IFInstruct comprises 10k samples, collected from multiple sources, including Visual Genome \cite{krishna2016visualgenomeconnectinglanguage}, LLaVA-Instruct-150k \cite{liu2023visualinstructiontuning}, and ALLaVA \cite{chen2024allavaharnessinggpt4vsynthesizeddata}, which provide a wide coverage of natural scenes, instruction-rich data, and diverse multimodal contexts. For the data example, you can refer to Figure \ref{fig:vc_instruct_example}.
\paragraph{VC-IFDPO}
To further investigate the impact of our vision-centric instruction-following data on improving model performance, we construct VC-IFDPO-10k, a large-scale dataset designed for preference learning. VC-IFDPO-10k is derived from our previously built VC-IFInstruct-10k dataset: for each image–question instance, we use the responses from VC-IFInstruct as the selected (preferred) answers. To construct the corresponding rejected answers, we adopt two complementary strategies: instruction ablation, where part of the constraints is deliberately removed or ignored, and visual edition, where the image input is edited to disrupt visual grounding. This design produces rich positive–negative preference pairs that enable vision-centric preference optimization through DPO training. Please refer to Figures \ref{fig:vc_dpo_example} and \ref{fig:dpo_rejected} for detailed data example.\\
\textbf{Instruction Ablation.}  
In this setting, we weaken the original multimodal instruction by removing a subset of its constraints. The selected answer strictly satisfies all constraints, while the rejected answer is generated under ablated instructions.\\
\textbf{Image Edit.}
In this setting, the instruction remains unchanged, but the image input is perturbed to disrupt visual grounding. For each question, we identify the key visual entities mentioned in the instruction, and apply visual edition by masking or editing these regions. The rejected answer comes from the perturbed input, thereby explicitly penalizing reliance on linguistic priors alone. More details about image edit could be found in Figure.\ref{fig:image edition}.
\subsection{Human Validation of Data Quality}
To evaluate the quality and consistency of our datasets, we conducted human annotation on a total of 400 samples, comprising 200 from VC-IFInstruct and 200 from VC-IFDPO.
Each instance was independently annotated by two trained annotators following standardized guidelines defining visual grounding, constraint satisfaction, and multimodal reasoning fidelity.

For VC-IFInstruct, annotators assessed whether each model-generated response satisfied the provided visual constraints, including spatial relations, attribute grounding, and comparative or counting requirements.
Each constraint was scored as either satisfied (1) or violated (0), and samples achieving at least 83\% constraint satisfaction were considered valid. Under this definition, 84\% of the VC-IFInstruct samples are judged as valid, indicating that the majority of constructed instruction–constraint pairs are well-formed and human-verifiable. For VC-IFDPO, annotators examined response pairs (chosen vs. rejected) to determine which response better aligned with the multimodal instruction and the corresponding visual content. We consider a preference pair valid if both annotators consistently prefer the chosen response over the rejected one. Using this strict criterion, 80\% of the VC-IFDPO pairs are validated by human annotators, suggesting that most preference annotations reflect meaningful and correctly oriented multimodal preferences.

We measure Inter-Annotator Agreement (IAA) \cite{cohen} to quantify annotation consistency across annotators.
The overall IAA score of 0.86 indicates a high level of agreement, demonstrating that our annotation process is both reliable and reproducible.
Please refer to Figure.\ref{fig:agreement} for information about human annotation.

\section{VC-IFEval}

We propose VC-IFEval, a challenging benchmark specifically designed to evaluate multimodal instruction-following with a strong emphasis on vision-centric capabilities. 
By systematically integrating diverse visual constraints such as spatial reasoning and attribute recognition, VC-IFEval enables a more fine-grained assessment of LVLMs' instruction-following ability beyond purely textual dimensions.
\subsection{VC-IFEval Construction}
VC-IFEval is constructed by generating image--task--constraint pairs using VC-IFEngine, followed by human refinement to resolve potential ambiguities.  We further apply a post-processing step using GPT-4o to identify and mitigate residual constraint conflicts. The final benchmark consists of 300 curated image--task--constraint pairs, covering diverse visual constraint types and image domains.

\subsection{Hybrid Evaluation}
Existing instruction-following benchmarks typically adopt a combination of rule-based matching and LLM-based judging, since their instructions are text-only and can be easily verified through predefined rules. However, this strategy becomes infeasible in our setting: the instructions in VC-IFEval are vision-centric, and many of the constraints (e.g., spatial relations, object attributes) cannot be evaluated by simple rule matching. At the same time, multimodal models often suffer from semantic priors, producing outputs that happen to follow the instructions while ignoring the visual input. To address these challenges, we design a hybrid evaluation framework that integrates two complementary components. On the one hand, we employ a powerful expert model (GPT-4o) to directly judge whether constraints are satisfied, providing a reliable measure of instruction adherence. On the other hand, we introduce a comparative evaluation strategy that compares model outputs generated with and without the image, thereby deriving an image influence signal that corrects for cases where models rely solely on linguistic priors without truly leveraging visual information. Further information about hybrid evaluation can be found in Figure \ref{fig:vc_ifeval_judge_templates}.

\paragraph{Direct Evaluation: MLLM-as-a-judge:} For direct evaluation, we adopt the MLLM-as-a-judge paradigm and employ GPT-4o as the expert model to assess constraint adherence. Given an image–task–constraint pair and a candidate response, GPT-4o is prompted to determine whether each constraint is satisfied. The judgments are aggregated into a binary score per instance, and the overall Constraint-following Accuracy (CFA) is computed as the proportion of satisfied constraints across the dataset. This provides a straightforward and standardized measure of instruction-following performance.
\paragraph{Comparative Evaluation:} While direct evaluation measures whether the model output satisfies the given constraints, such absolute scoring may be affected by semantic priors and linguistic biases. To mitigate this, we introduce comparative evaluation, which explicitly compares paired outputs generated with and without the image. Instead of independently judging each output, the evaluator is prompted to decide whether the inclusion of visual input leads to better constraint satisfaction. In this way, comparative evaluation provides a more reliable assessment of the image influence by isolating the contribution of visual grounding from purely textual reasoning.

\paragraph{Score Formulation:} We integrate the outcomes of direct and comparative evaluation into a unified scoring framework. 
Let $N$ denote the total number of evaluation instances. 
For each instance $i$, we obtain two components: (1) the Constraint-following Accuracy (CFA), which indicates whether the response satisfies the given constraints under direct evaluation, and (2) the Image Influence Score (IIS), which reflects whether the response generated with the image is preferred over that generated without the image under comparative evaluation. 
The final hybrid score of a model is defined as
\[
S_{\text{model}} = \frac{1}{2N} \sum_{i=1}^{N} \big( \text{CFA}_i + \text{IIS}_i \big),
\]
which balances constraint adherence with effective utilization of visual information, providing a holistic measure of multimodal instruction-following capability.

\subsection{Evaluation Robustness Validation}
To further validate the reliability of our automatic evaluation protocol, we conduct a human agreement study on a randomly sampled subset of 400 instances from the VC-IFDPO dataset. Human annotators are asked to assess the correctness of the evaluation outcomes produced by our protocol under both direct and comparative judging settings.

For direct judging, we focus on negative samples and evaluate agreement at the constraint level. Human annotators examine whether each individual visual constraint is indeed violated as indicated by the automatic evaluator. The resulting agreement rate reaches 90\%, suggesting that the direct judgment of constraint violation or satisfaction aligns well with human assessments.

For comparative judging, we generate additional responses for all samples by removing the image input and ask both the automatic evaluator and human annotators to compare responses with and without visual input. The agreement rate between human judgments and automatic comparative evaluation reaches 92\%, indicating that the evaluator consistently captures whether model outputs genuinely depend on visual information.

We additionally examine whether the evaluation outcomes are stable across different judging models. Besides the primary GPT-based evaluator, we employ the open-source Qwen2.5-VL model as an alternative evaluator and compare their direct judgments on the same evaluation instances. We observe a high inter-model agreement of 86\%, suggesting that the evaluation results are largely consistent across evaluators and are not artifacts of a particular proprietary judging model. This consistency also indicates the feasibility of using open-source evaluators to reduce evaluation cost while maintaining reliable assessment.


\section{Experiments}
\subsection{Experiment Setup}
We evaluate our models on a series of instruction-following and visual reasoning benchmarks, including MM-IFEval \cite{ding2025mm}, VC-IFEval, and IFEval \cite{zhou2023instruction}, where IFEval is language-only while the others are multimodal. We further assess their generalization on several VQA benchmarks such as [list here]. All experiments are conducted on H20 96 GB GPUs using two representative multimodal large language models, Qwen2.5-VL-7B-Instruct \cite{bai2025qwen25vltechnicalreport} and LLaVA-NeXT-Llama3-8B \cite{liu2024llavanext}. Training and evaluation are implemented with the LLaMA-Factory \cite{zheng2024llamafactory} framework and VLMEvalKit \cite{duan2024vlmevalkit}, respectively, using a batch size of 8 and a learning rate of 1e-5 for both SFT and DPO stages. 
\subsection{Main Results about VC-IFInstruct and VC-IFDPO}
\begin{table*}[t]
  \centering
  \small
  \begin{tabular}{lccccc}
    \hline
    \textbf{Model} & \textbf{VC-IFEval (Ours)} & \textbf{MM-IFEval} & \multicolumn{3}{c}{\textbf{IFEval}} \\
    \cline{4-6}
                   &                           &                    & \textbf{Prompt} & \textbf{Instruction} & \textbf{Avg.} \\
    \hline
    Qwen2.5-VL-7B-Instruct     & 57.3 & 54.3  & 64.5 & 74.2 & 69.4 \\
    w. VC-IFInstruct-10k  & 63.0 & 55.0 & 64.1 & 75.0 & 69.6 \\
    w. VC-IFDPO-10k &66.1 & 55.8&65.0 &75.2 & 70.1\\
    \hline
    LLaVA-NeXT-Llama3-8B           & 50.1 & 50.2 & 45.8 & 57.7 & 51.8 \\
    w. VC-IFInstruct-10k          & 53.3 & 50.9 & 47.5 & 59.0 & 53.3 \\
    w. VC-IFDPO-10k & 60.2 & 52.1&46.8 &61.2 & 54.0\\
    \hline
  \end{tabular}
  \caption{\label{tab:eval-comparison}
    Comparison across instruction-following benchmarks.
  }
\end{table*}

\begin{table*}[t]
  \centering
  \small
  \begin{tabular}{lccccc}
    \hline
    \textbf{Model}  & \textbf{MMT-Bench} & \textbf{MMBench} & \textbf{MMVet} & \textbf{POPE} \\
    \hline
    Qwen2.5-VL-7B-Instruct       & 67.8 & 83.2 & 67.2 & 88.3 \\
    w. VC-IFInstruct-10k          & 66.5 & 82.1 & 65.6 & 88.7 \\
    w. VC-IFDPO-10k               & 67.4 & 82.8 & 68.3 & 88.5 \\
    \hline
    LLaVA-NeXT-Llama3-8B        & 53.1 & 72.5 & 43.3 & 87.2 \\
    w. VC-IFInstruct-10k          & 54.3 & 73.2 & 44.0 & 87.0 \\
    w. VC-IFDPO-10k               & 53.2 & 73.6 & 44.2 & 86.8 \\
    \hline
  \end{tabular}
  \caption{\label{tab:vqa}
    Comparison across VQA benchmarks (MMT-Bench\cite{mmt-bench}, MMBench\cite{mmbench}, MMVet\cite{mmvet}, POPE\cite{POPE}).
  }
\end{table*}

\begin{table}[t]
  \centering
  \small
  \begin{tabular}{lc}
    \hline
    \textbf{Model} & \textbf{VC-IFEval} \\
    \hline
    Qwen2.5-VL-7B-Instruct     & 57.3 \\
    w.VC-IFInstruct  & 63.0 \\
    w.VC-IFDPO           & 66.1 \\
    w.MM-IFInstruct          & 60.1 \\
    w.MM-IFDPO          & 61.2\\
    \hline
  \end{tabular}
  \caption{\label{tab:vc-ifeval}
 Performance comparison of models trained on our datasets and datasets by MM-IFEngine \cite{ding2025mm}.
  }
\end{table}

\paragraph{Results on Instruction-Following Benchmarks}
As shown in Table~\ref{tab:eval-comparison}, both Qwen2.5-VL-7B-Instruct and LLaVA-NeXT-Llama3-8B exhibit substantial performance enhancements after being trained on our proposed VC-IFInstruct and VC-IFDPO datasets. 
Training on VC-IFInstruct leads to a significant improvement on visual instruction-following benchmarks, with an average increase of 4.45\%. Moreover, the models also achieve consistent gains on textual instruction-following benchmarks, with an average improvement of 0.7\% on MM-IFEval and 0.85\% on IFEval. Similarly, fine-tuning with VC-IFDPO further enhances model performance, yielding an average gain of 9.45\% on visual benchmarks and improvements of 1.7\% and 1.45\% on MM-IFEval and IFEval, respectively. These results demonstrate that our visual-centric datasets not only strengthen the model’s ability to follow visually grounded instructions but also generalize effectively to textual benchmarks, reflecting the high quality and broad generalization capability of our data.
\paragraph{Results on VQA Benchmarks}
We further evaluate the models trained on VC-IFInstruct and VC-IFDPO on several visual question answering (VQA) benchmarks to examine their generalization to visual reasoning tasks. As shown in Table~\ref{tab:vqa}, both Qwen2.5-VL-7B-Instruct and LLaVA-NeXT-Llama3-8B demonstrate consistent performance after visual-centric instruction tuning. These results verify that our visual-centric instruction data enhance not only instruction-following ability but also general visual reasoning capability, underscoring the broader applicability of our approach.
\paragraph{Comparison with MM-IFInstruct and MM-IFDPO}
As shown in Table~\ref{tab:vc-ifeval}, models trained with our visual-centric datasets achieve better performance on visual-centric instruction-following benchmarks than those trained with the MM-IFEngine datasets. Compared to MM-IFInstruct, training on VC-IFInstruct leads to a clear improvement on VC-IFEval, demonstrating that incorporating visual-centric instructions encourages models to better ground textual reasoning in visual context. Further applying DPO training on VC-IFDPO brings additional gains, outperforming MM-IFDPO by a noticeable margin. This result indicates that our visual-centric preference optimization more effectively aligns the policy with multimodal feedback, leading to more accurate and visually coherent responses. Overall, these comparisons confirm that our datasets provide higher-quality alignment signals and foster a more faithful understanding of vision-conditioned instructions than those from MM-IFEngine.
\paragraph{SFT vs DPO}
As evidenced by Table~\ref{tab:eval-comparison} and Table~\ref{tab:vqa}, DPO training consistently outperforms SFT, achieving higher accuracy on both instruction-following benchmarks and VQA benchmarks. This improvement can be attributed to the additional contrastive supervision introduced in DPO, which helps the model better distinguish between preferred and non-preferred responses, thereby enhancing instruction-following capability.
\subsection{Ablation Studies}
\paragraph{Ablation Studies on DPO Settings}
As shown in Table~\ref{tab:dpo-settings}, we conduct ablation studies on various strategies for constructing DPO data. The first three settings remove different proportions of visual constraints (-33\%, -66\%, and -100\%) to generate rejected responses, while the last two settings focus on the impact of visual information: using edited images for 50\% of the samples and removing visual inputs for the remaining 50\% (image edit -50\%), and completely removing all visual inputs (w/o image). We observe that DPO with complete constraint removal (-100\%) achieves the highest performance, while removing all image inputs results in weaker instruction-following ability. 
This comparison suggests that visual editing is more effective than omitting visual inputs, emphasizing the importance of maintaining visual grounding during DPO data construction.

\begin{table}[t]
  \centering
  \small
  \begin{tabular}{lcc}
    \hline
    \textbf{Model / Setting} &  \textbf{VC-IFEval} \\
    \hline
    Qwen2.5-VL-7B-Instruct          & 57.3\\
    + DPO (-33\% cons)              &  64.4\\
    + DPO (-66\% cons)              & 67.3 \\
    + DPO (-100\% cons)             & 68.8 \\
    + DPO (w/o image)               & 59.3 \\
    + DPO (image edit -50\%)     & 60.6 \\
    \hline
  \end{tabular}
  \caption{\label{tab:dpo-settings}
Comparison between different DPO settings.
  }
\end{table}
\section{Conclusion}
This paper advances multimodal instruction-following by introducing VC-IFEngine, a visual-centric pipeline for constructing instruction-tuning data. Using this framework, we build two datasets—VC-IFInstruct-10k for supervised fine-tuning and VC-IFDPO-10k for preference optimization—and propose VC-IFEval, a benchmark designed to evaluate models’ capability to follow vision-conditioned instructions. Experiments demonstrate that our visual-centric data lead to significant improvements in visual and textual instruction-following performance, highlighting the importance of explicitly modeling visual constraints in data construction. We hope this work serves as a step toward deepening the understanding of multimodal instruction following and accelerating the reliable adoption of multimodal large language models in real-world applications.

\section*{Limitations}

Despite the effectiveness of our proposed framework, several limitations remain to be addressed in future work.
First, our evaluation pipeline heavily relies on GPT-4o as the judging model, which introduces both cost and reliability concerns.
Due to the complexity of visual instruction-following tasks—where constraints are dynamically generated based on both image content and task semantics—rule-based or automated evaluation methods are difficult to apply.
As a result, we depend on expert models for consistent and high-quality judgments.
We acknowledge that this dependency increases computational overhead and may introduce bias from the evaluator model itself.
Future work should explore more efficient and objective evaluation mechanisms that reduce cost while maintaining reliability.

While our visual constraints effectively encourage models to focus on vision-grounded instructions, an excessive number of constraints may overly restrict the model’s expressive ability.
Compared with textual instructions, visual instructions rely heavily on the richness of the image content.
For images with relatively simple or less informative scenes, enforcing too many constraints can lead to redundancy, limited diversity, and inefficient training.
Future improvements should consider adaptive constraint allocation based on image complexity to balance visual grounding and generative flexibility.

Finally, we leave the exploration of real-world applications of vision-centric instruction-following to future research.
Our ultimate goal is to enhance the trustworthiness of Multimodal Large Language Models (MLLMs) and to promote their safe and reliable deployment across broader domains such as education, healthcare, and human–AI interaction.

\bibliography{acl_latex}

\clearpage
\appendix

\section{Case Study}
\label{sec:appendix}

\subsection{Detailed Limitation of MM-IFEval}
\begin{figure*}[t]
\centering
\begin{tcolorbox}[
    colback=blue!1!white,
    colframe=blue!70!black,
    title=MM-IFEval Limitation: Constraint Following Without Visual Input,
    colbacktitle=blue!5!white,
    coltitle=blue!50!black,
    fonttitle=\bfseries,
    boxrule=0.5pt,
    arc=4pt,
    boxsep=5pt,
    left=6pt,
    right=6pt,
    top=6pt,
    bottom=6pt
]
\begin{minipage}[t]{0.42\linewidth}
    \centering
    \includegraphics[width=\linewidth]{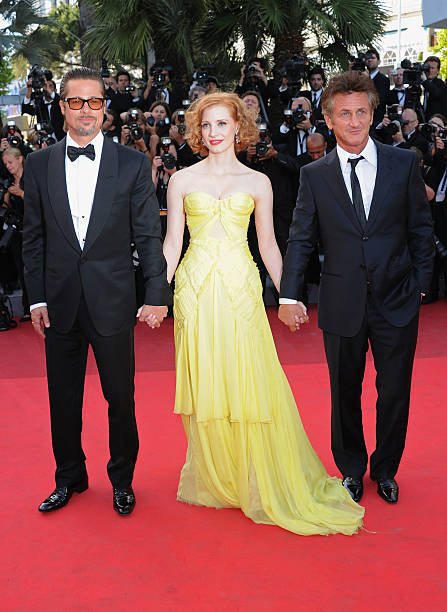}
    \vspace{3pt}
    {\small \textit{(a) Image.}}\\[6pt]

    \raggedright
    \footnotesize
    \textbf{(b) Task \& Constraints.}\\
    Task: Imagine you are the photographer who took this picture. Briefly explain why you chose to capture this particular moment and what story you hope it conveys.\\
    Constraints: Present your explanation in two main points: (1) Why you chose to capture this moment, and (2) The story you hope it conveys. Use a brief bulleted sub-list under each point to detail specific elements or emotions. Provide a concise title that summarizes the main idea, such as \emph{Capturing the Essence of Everyday Life} or \emph{A Moment Frozen in Time}. Your response must consist of exactly three paragraphs: one for the introduction, one for the main explanation, and one for the conclusion. Conclude with a final paragraph summarizing the key arguments and reiterating the emotional impact of the photograph. Compose the response as if it were a personal journal entry, reflecting on the moment and the emotions it evokes.
\end{minipage}
\hfill
\begin{minipage}[t]{0.55\linewidth}
    \raggedright
    \small
    \textbf{(c) Model Response (without visual input).}\\
    \textit{Journal Entry: "The Quiet Beauty of Stillness"}\\
    \textbf{Introduction.}  
    Today, as I wandered through the city streets, something caught my eye—a fleeting moment of stillness amidst the chaos. The scene was simple yet profound, and it felt like a whisper from the universe, urging me to capture its essence. As a photographer, I often find myself drawn to moments that transcend the ordinary, offering glimpses into the deeper layers of life. This particular image, though seemingly empty at first glance, holds a story waiting to be told.\\[3pt]
    \textbf{Main Explanation.}  
    I chose to capture this moment because it offered contrast with chaos—the surrounding streets were alive with sound and motion, yet here was a pause, a fragment of silence. The emptiness carried emotion, a quiet vulnerability that made me reflect on how fleeting and meaningful stillness can be. The composition, marked by simplicity, drew the eye to open space and invited interpretation.\\[3pt]
    \textbf{Conclusion.}  
    The photograph is not about what is visible but what lies beneath: emotion, memory, and introspection. By freezing this still moment, I sought to remind both myself and others of the quiet power in ordinary scenes. In a world that moves too fast, this image invites us to pause, breathe, and find beauty in stillness.\\[6pt]

    \textbf{Observation.}\\
    Even without visual input, the model produces a response that fully satisfies the textual constraint, demonstrating reliance on \emph{semantic priors} rather than visual grounding. This example highlights a core limitation of MM-IFEval, where textual compliance can be mistaken for multimodal understanding.
\end{minipage}
\end{tcolorbox}

\caption{
An example of MM-IFEval limitation, where the model follows textual constraints without visual grounding.
}
\label{fig:mmifeval_limitation}
\end{figure*}

Although MM-IFEval extends the instruction-following paradigm into the multimodal domain, its underlying evaluation mechanism remains largely text-oriented. 
In many cases, multimodal large language models (MLLMs) can still rely on textual priors to follow instructions even without referring to the visual input. 
As a result, responses that conform to the textual component of an instruction but contradict the actual visual content may still be judged as correct by the model-based evaluator. 
Such cases illustrate that MM-IFEval does not fully exploit the multimodal information available in the input, leading to potential bias in both dataset construction and evaluation methodology. 
Representative examples of this phenomenon are shown in Figure~\ref{fig:mmifeval_limitation}. 
\section{VC-IFEngine}
\subsection{Image Filter and Sources}
We collect images from three major instruction-following datasets—Allava \cite{chen2024allavaharnessinggpt4vsynthesizeddata}, Visual Genome \cite{krishna2016visualgenomeconnectinglanguage}, and LLaVA-Instruct-150k \cite{liu2023visualinstructiontuning}—covering both natural and synthetic scenes. Allava provides diverse real-world photographs with rich object co-occurrence and attribute variations. Visual Genome contributes densely annotated scenes with explicit object–relationship graphs and bounding boxes, supporting high-quality constraint attachment. LLaVA-Instruct-150k offers instruction-aligned images that ensure contextual correspondence between visual content and language tasks. 

\begin{figure}[t]
    \centering
    \includegraphics[width=\linewidth]{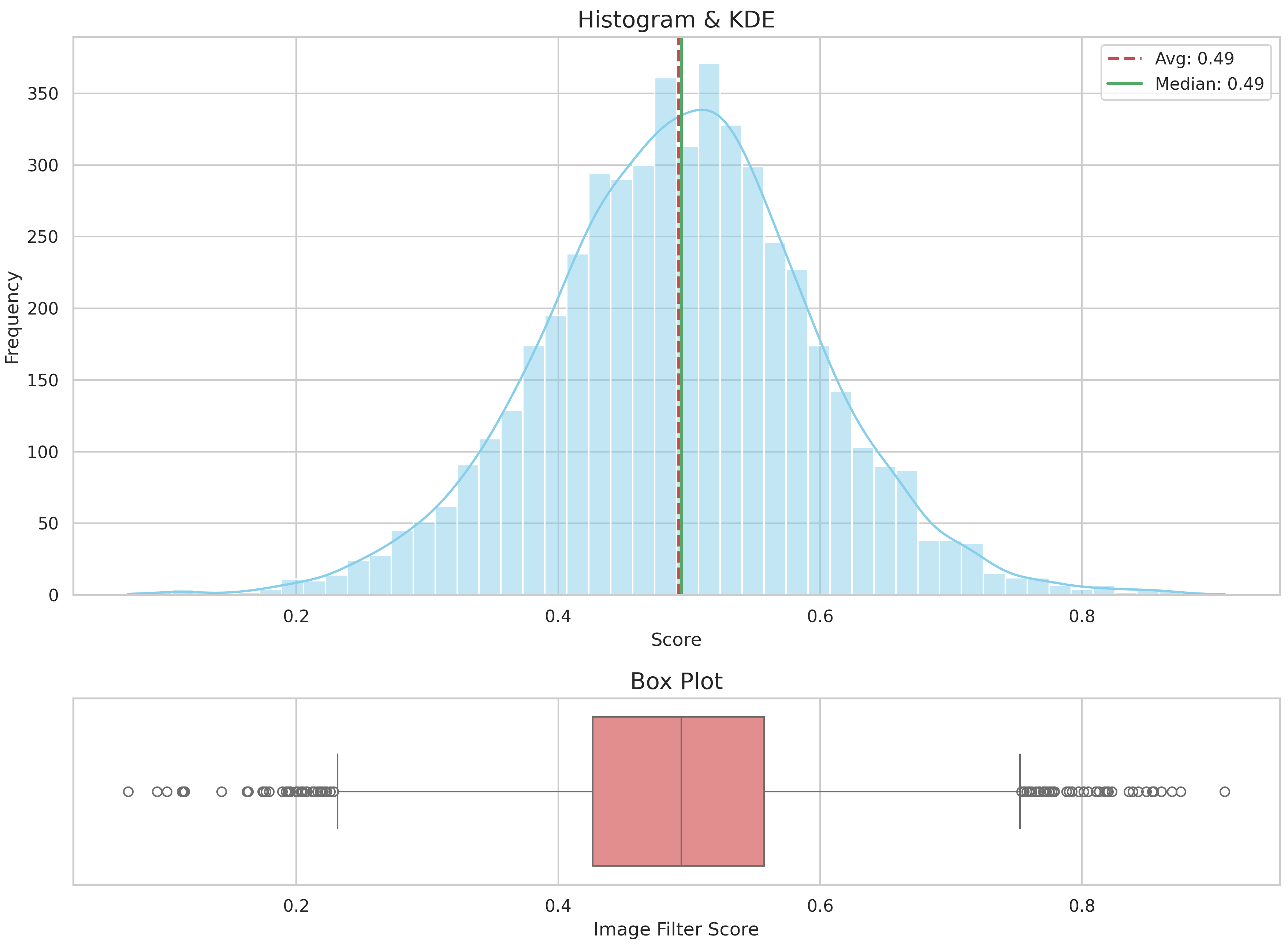}
    \caption{Distribution of IC9600 information-density scores across sampled images from Allava, Visual Genome, and LLaVA-Instruct-150k.}
    \label{fig:image_filter_dist}
\end{figure}

To further validate the filtering process, we randomly sample a subset of images from all sources and visualize the distribution of IC9600 \cite{feng2023ic9600} scores, as shown in Figure~\ref{fig:image_filter_dist}. 
Both the mean and median scores are approximately 0.50, indicating a near-symmetric distribution of visual information density across the image pool. 
Based on this observation, we set 0.50 as the filtering threshold: images with IC9600 scores above this value are retained for task and instruction generation, while those below are discarded. 
\subsection{Task and Constraint Generation Details}
\label{appendix:task_constraintgeneration}
For image sources that do not include original task instructions or task instructions that are not suitable for visual development, we construct a diverse external task list designed to elicit open-ended, semantically rich responses from multimodal models. 
Each image is paired with one template sampled from this pool, which guide the model to describe, reason, or analyze the visual content from different perspectives. 
In total, we define 18 distinct task templates, grouped into five major categories: Descriptive \& Analytical, Reasoning \& Interpretive, Emotional \& Perspective, Creative \& Narrative, and Instructional \& Role-based. The specific categories and examples are shown in Table~\ref{tab:task_list}.

Following task generation, we construct a constraint pool to further enhance the visual dependency of each instruction. Constraints are formulated as auxiliary conditions that guide the model to ground its responses in specific visual evidence, ensuring that the task cannot be completed solely through textual reasoning. 
We divide the constraint pool into two major categories—Objective \& Analytical and Inferential \& Creative—covering a total of ten representative constraint types. 
The constraint types are listed below, and more detailed information and examples can be found in Table~\ref{tab:constraint_pool}.

\paragraph{Category 1: Objective \& Analytical} These constraints focus on accurately extracting and analyzing visual information directly observable in the image.
\begin{itemize}[leftmargin=1.5em,itemsep=2pt,topsep=2pt]
  \item \textbf{Spatial:} Emphasizes object positions, layouts, and relative arrangements within the scene.
  \item \textbf{Attribute:} Focuses on visual properties such as color, texture, material, or physical state.
  \item \textbf{Comparative:} Requires comparison between multiple objects or regions within the image.
  \item \textbf{Counting \& Numerical:} Involves precise object counting or quantitative reasoning about visual elements.
  \item \textbf{Textual (OCR) \& Grounding:} Targets reading, transcribing, or grounding textual information present in the image.
\end{itemize}

\paragraph{Category 2: Inferential \& Creative} These constraints require reasoning beyond literal perception, involving inference, imagination, and conceptual understanding.
\begin{itemize}[leftmargin=1.5em,itemsep=2pt,topsep=2pt]
  \item \textbf{Causal \& Temporal:} Focuses on reasoning about event order, causality, or temporal dynamics implied by the scene.
  \item \textbf{Affective \& Atmospheric:} Highlights the emotional tone, mood, or atmosphere conveyed by the visual content.
  \item \textbf{Perspective \& Role-Play:} Alters the viewpoint or role from which the image should be interpreted.
  \item \textbf{Hypothetical \& Counterfactual:} Encourages imagination of how the scene would change under alternative or hypothetical conditions.
  \item \textbf{Abstract \& Conceptualization:} Connects concrete visual details to abstract ideas, metaphors, or high-level concepts.
\end{itemize}

\subsection{Prompt Templates in Data Generation}
VC-IFEngine provides a scalable and modular pipeline for constructing visual-centric instruction-following datasets. 
It supports both images with pre-existing instructions and those without any textual annotations, enabling systematic generation of multimodal tasks at scale. 
To ensure linguistic diversity and consistency across data samples, we employ a series of carefully designed prompt templates that guide large language models through two key stages of generation: the instruction generation and the constraint generation. These templates standardize the data construction process while maintaining sufficient variability across task types and visual domains. Representative prompt templates from both modules are illustrated in Figures~\ref{fig:task_prompt_templates} and~\ref{fig:constraint_prompt_templates}, showing how VC-IFEngine transforms raw images into structured, visually grounded instruction–response pairs.

\subsection{VC-IFInstruct and VC-IFDPO}
Our VC-IFInstruct dataset is constructed using the VC-IFEngine pipeline based on filtered images from Allava, Visual Genome, and LLaVA-Instruct-150k. 
Each image is paired with an instruction generated from the task and constraint pools, resulting in a large-scale collection of vision-dependent instruction–response pairs. 
These samples cover a wide range of visual reasoning and compositional understanding scenarios, ensuring that the generated data strongly emphasizes visual grounding.

\begin{figure*}[t]
\centering
\begin{tcolorbox}[
    colback=blue!1!white,
    colframe=blue!70!black,
    title=VC-IFInstruct Example,
    colbacktitle=blue!5!white,
    coltitle=blue!50!black,
    fonttitle=\bfseries,
    boxrule=0.5pt,
    arc=4pt,
    boxsep=5pt,
    left=6pt,
    right=6pt,
    top=6pt,
    bottom=6pt,
    breakable
]
\small

\noindent
\begin{minipage}[t]{0.44\linewidth}
    \centering
    \vspace{0pt} 
    \includegraphics[width=\linewidth]{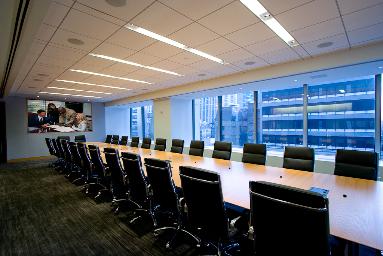}
    \vspace{3pt}
    {\small \textit{(a) Image.}}
\end{minipage}
\hfill
\begin{minipage}[t]{0.53\linewidth}
    \vspace{0pt} 
    \raggedright
    \small
    \textbf{(b) Prompt.}\\
    Describe the modern conference room featuring a long rectangular table surrounded by black office chairs. Focus on the spatial arrangement of the chairs, which are symmetrically aligned on both sides, reflecting a collaborative and orderly setting. Note the contrast between the dark, soft carpet and the polished furniture surfaces, emphasizing the balance between warmth and professionalism. Compare the evenly distributed artificial ceiling lights with the natural daylight from the large windows, analyzing how the combination enhances the room’s brightness and openness. Count the total number of chairs and mention their consistent design. Briefly interpret the content on the background screen and infer the possible meeting topic. Finally, predict how the atmosphere would change if the blinds were closed or the lighting dimmed, shifting from a bright and open environment to a more intimate and focused one.
\end{minipage}

\vspace{8pt}

\raggedright
\small
\textbf{(c) Model Response.}\\
The image depicts a modern conference room designed for professional meetings, characterized by a sleek and minimalist aesthetic. Here's a detailed analysis:

\textbf{Key Elements of the Room:}

1. \textbf{Conference Table:}
   The room features a long rectangular conference table, which is the focal point of the space. Its clean, polished surface suggests a professional and functional design.
   The table's length and central positioning allow for ample seating along both sides, fostering an inclusive and collaborative environment.

2. \textbf{Chairs:}
   There are at least 16 black office chairs, all uniformly positioned along the table. The chairs are modern, with a sleek design featuring high backs and adjustable armrests, indicating comfort and ergonomic considerations.
   The chairs' arrangement is symmetrical, enhancing the room's sense of order and equality among participants, as everyone has equal access to the table and screen.

3. \textbf{Carpeting:}
   The carpet is a dark-colored, soft fabric that contrasts with the otherwise hard and polished surfaces of the furniture and walls. This choice adds warmth and a touch of comfort to the space, balancing the room's minimalism with a subtle humanizing element.

4. \textbf{Lighting:}
   The room is well-lit, with a combination of artificial lighting from recessed ceiling panels and natural light streaming through large windows.
   The ceiling lights are bright and evenly distributed, ensuring good visibility for all participants. The natural light from the windows enhances the room's spaciousness and creates a welcoming, open atmosphere.
   The interplay between artificial and natural light contributes to a balanced and inviting environment, suitable for extended meetings or presentations.

5. \textbf{Windows:}
   The large windows on one side of the room offer a view of the cityscape, suggesting that the room is located in a high-rise building. This view not only adds aesthetic value but also contributes to a sense of openness and connection to the outside world.
   The windows allow for ample natural light, which can boost mood and productivity.

6. \textbf{Screen and Presentation:}
   A screen in the background displays what appears to be a presentation or discussion topic. The content is not entirely clear, but it suggests that the room is set up for a meeting, workshop, or collaborative session.
   The presence of the screen indicates a focus on information-sharing or decision-making, reinforcing the professional nature of the space.

\textbf{Spatial Arrangement:}
The uniform placement of chairs along both sides of the table suggests a collaborative and inclusive meeting dynamic. Participants can easily engage with one another, fostering open communication and shared focus on the screen or discussion topic.
The long table design allows for a large group to be accommodated, indicating the room is suitable for meetings involving multiple stakeholders or departments.

\textbf{Atmosphere and Mood:}
The current atmosphere is bright, open, and professional. The combination of natural and artificial lighting creates a welcoming environment conducive to productive discussions.
If the blinds were closed or the lighting were dimmed, the atmosphere would shift significantly:
Closing the blinds would reduce natural light, making the room darker and more enclosed. This could create a more intimate or focused setting, perhaps suitable for brainstorming sessions or confidential discussions.
Dimming the artificial lighting would further enhance this effect, making the room feel more subdued and reflective. The mood would likely shift from open and energetic to more contemplative and serious.

\textbf{Summary:}
The conference room is a well-designed, professional space optimized for collaboration and productivity. The combination of modern furniture, a spacious layout, and balanced lighting creates an inviting environment. The symmetrical arrangement of chairs and the presence of a presentation screen suggest a focus on inclusivity and information-sharing. The room's atmosphere is currently bright and open, but could be adjusted to suit different meeting needs by manipulating the lighting and window coverings. Overall, the space is versatile and well-suited for a variety of professional settings.

\end{tcolorbox}

\caption{
An example of VC-IFInstruct data.
}
\label{fig:vc_instruct_example}
\end{figure*}

\begin{figure*}[t]
\centering
\begin{tcolorbox}[
    colback=blue!1!white,
    colframe=blue!70!black,
    title=VC-IFDPO Example,
    colbacktitle=blue!5!white,
    coltitle=blue!50!black,
    fonttitle=\bfseries,
    boxrule=0.5pt,
    arc=4pt,
    boxsep=5pt,
    left=6pt,
    right=6pt,
    top=6pt,
    bottom=6pt,
    breakable
]
\small

\noindent
\begin{minipage}[t]{0.44\linewidth}
    \centering
    \vspace{0pt}
    \includegraphics[width=\linewidth]{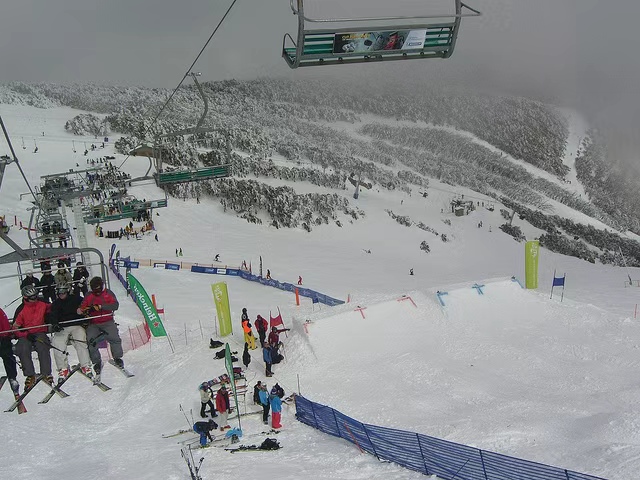}
    \vspace{3pt}
    {\small \textit{(a) Image.}}
\end{minipage}
\hfill
\begin{minipage}[t]{0.53\linewidth}
    \vspace{0pt}
    \raggedright
    \small
    \textbf{(b) Prompt}\\[3pt]
    Describe the picture, focusing on the spatial arrangement of people, ski lifts, and snow-covered terrain. Analyze how individuals in the foreground, such as those near the ski lifts, are engaged in activities like preparing to ski or waiting in line, while those in the background are spread across the slopes, either skiing or observing. Compare the density of people near the ski lifts, which appears higher due to the concentration of activity, with the more dispersed groups on the slopes. Predict how the flow of movement and activity might change if the ski lift were to stop, potentially causing a backlog near the lift, or if more people arrived, increasing congestion and activity levels. Consider the overall mood and atmosphere, noting the overcast weather conditions that contribute to a calm yet active ambiance, with spectators adding to the sense of a lively event. Finally, imagine how the scene might evolve if the event were to shift to a different time of day, such as evening, when the lighting would create a more dramatic and possibly less crowded atmosphere, or to a different season, where the absence of snow might transform the setting entirely.
\end{minipage}

\vspace{8pt}

\raggedright
\small
\textbf{(c) Chosen}\\

This image captures a lively ski resort scene, showing a dynamic interaction between people, infrastructure, and the surrounding natural landscape. The spatial arrangement of the scene can be understood by examining how people are distributed across different regions of the image.

In the foreground, activity is concentrated around the ski lifts. Several individuals are preparing to board the lift, adjusting their equipment, or standing in a queue. Because the ski lift serves as a central piece of infrastructure, the density of people in this area is noticeably higher. Most individuals are engaged in skiing related tasks such as tying boots, adjusting ski poles, or waiting for their turn, making this region the primary focal point of movement and preparation.

The midground features a snow covered slope with a mix of skiers and snowboarders navigating the terrain. Some are skiing downhill, while others appear to be preparing to descend or taking short breaks. Flags and barriers mark designated skiing areas, possibly indicating regulated zones or competitive sections. Compared to the foreground, people in the midground are more evenly distributed, resulting in a less crowded appearance.

In the background, the scene opens into an expansive snow covered landscape with slopes extending into the distance. Trees and terrain are blanketed in snow, and only a small number of individuals are visible at this depth. The decreasing density of people emphasizes the scale and openness of the environment.

Overall, the highest concentration of people appears near the ski lifts, while the slopes show a more dispersed pattern of movement. If the ski lift were to stop operating, congestion would likely increase in the foreground as skiers accumulate while waiting to ascend. Additional arrivals would further intensify crowding near the lift area, leading to longer waiting times and reduced mobility.

The overcast weather contributes to a calm yet active atmosphere. Diffuse lighting softens the scene and reduces strong shadows, creating a visually cohesive appearance. The presence of spectators suggests that the resort may be hosting a popular event or experiencing high attendance. Taken together, the image effectively captures the contrast between concentrated activity near key infrastructure and freer movement across the open slopes, forming a balanced and dynamic winter scene.
\\

\end{tcolorbox}

\caption{
An example of VC-IFDPO data
}
\label{fig:vc_dpo_example}
\end{figure*}

\begin{figure*}[t]
\centering
\begin{tcolorbox}[
    colback=blue!1!white,
    colframe=blue!70!black,
    title=Image Edition Example,
    colbacktitle=blue!5!white,
    coltitle=blue!50!black,
    fonttitle=\bfseries,
    boxrule=0.5pt,
    arc=4pt,
    boxsep=5pt,
    left=6pt,
    right=6pt,
    top=6pt,
    bottom=6pt,
    breakable
]
\small

\noindent
\begin{minipage}[t]{0.47\linewidth}
    \centering
    \vspace{0pt}
    \includegraphics[width=\linewidth]{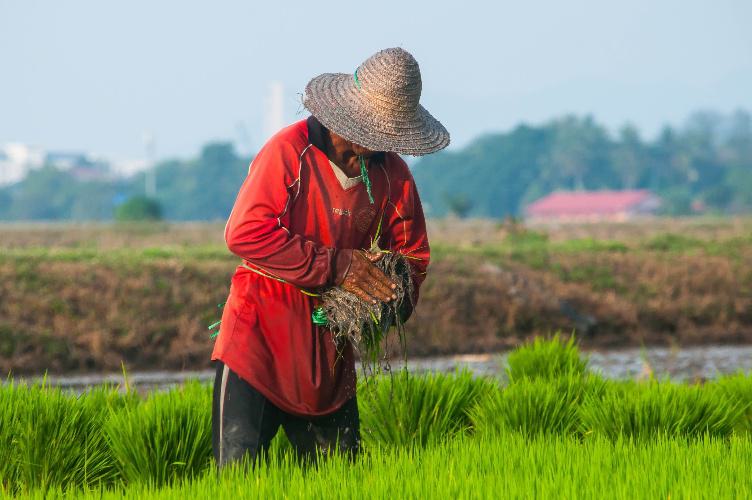}
    \vspace{3pt}
    {\small \textit{(a) Original Image.}}
\end{minipage}
\hfill
\begin{minipage}[t]{0.47\linewidth}
    \centering
    \vspace{0pt}
    \includegraphics[width=\linewidth]{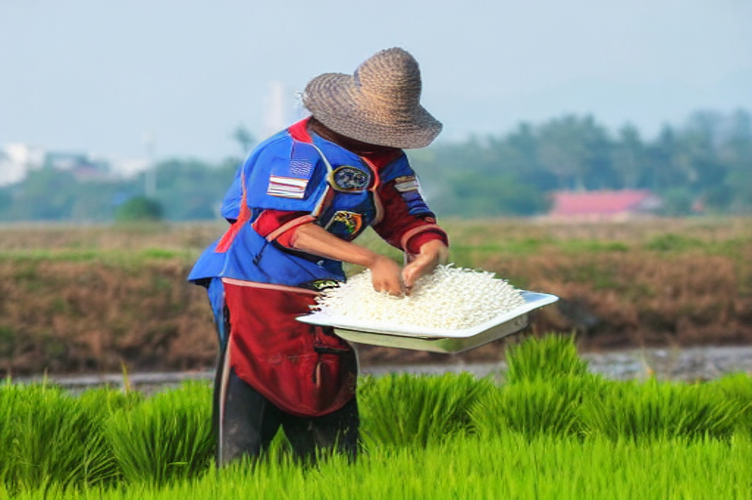}
    \vspace{3pt}
    {\small \textit{(b) Edited Image.}}
\end{minipage}

\vspace{10pt}
\textbf{Image Edit Details}
\small In the image edit setting, half of the filtered images are randomly selected for image editing. The edited images are then evaluated by an expert model to assess the quality and success of the modifications. Based on this evaluation, the final DPO dataset includes 10K samples, of which 20\% correspond to the image edition setting. The modification success rate for the image edition process is 46.2\%.
\vspace{4pt}
\raggedright
\small

\end{tcolorbox}

\caption{
An example of the image editing setting used in VC-IFDPO. We employ Stable Diffusion \cite{rombach2021highresolution} to modify the salient regions of the image while preserving the overall visual context, creating visually coherent yet semantically perturbed scenes that induce controlled visual ambiguity.
}
\label{fig:image edition}
\end{figure*}

To build the VC-IFDPO dataset for preference optimization, we extend VC-IFInstruct by introducing contrastive pairs composed of preferred and rejected responses. 
Specifically, 80\% of the DPO pairs are generated through constraint ablation, where 50\% of the visual constraints are randomly removed to create rejected responses, while the remaining 20\% are constructed using image edition, in which visual inputs are modified to introduce subtle inconsistencies. Representative samples from VC-IFInstruct and VC-IFDPO are shown in Figures~\ref{fig:vc_instruct_example} and~\ref{fig:vc_dpo_example},\ref{fig:dpo_rejected}, respectively, illustrating generated instructions and DPO preference pairs. Figure~\ref{fig:image edition} further visualizes the image edition strategy used to create visually influenced inputs in VC-IFDPO.
\section{VC-IFEval}
\subsection{Direct Judge and Comparative Judge}
To evaluate constraint adherence while controlling for semantic priors, we run two judging protocols: Direct Judge and Comparative Judge. The Direct Judge inspects a single response given the instruction and constraint and decides whether the constraint is satisfied. The Comparative Judge isolates image influence by eliciting two responses from the same model under identical textual prompts—one with the image and one without the image—and comparing them to verify that the visual input induces a consistent, constraint-relevant change in behavior. The unified prompt templates for both protocols are shown in Figure~\ref{fig:vc_ifeval_judge_templates}.

\begin{table*}[t]
\centering
\small
\setlength{\tabcolsep}{4pt}
\begin{tabular}{p{0.18\linewidth} p{0.25\linewidth} p{0.50\linewidth}}
\hline
\textbf{Category} & \textbf{ } & \textbf{Example Template} \\
\hline
\multicolumn{3}{l}{\textbf{Category 1: Descriptive \& Analytical}} \\
\hline
Scene Description & Describe the visual scene in detail, focusing on spatial arrangement and key elements. & “Describe all visible objects in the image and explain their relative positions.” \\
Object Analysis & Identify and explain the attributes or states of major objects in the image. & “Describe the texture, color, and condition of the main object.” \\
Comparison & Compare two or more visual elements and explain their differences. & “Compare the two people shown in the image in terms of clothing and posture.” \\
Counting & Count specific objects or visual entities that meet given criteria. & “Count the number of chairs visible in the scene.” \\
\hline
\multicolumn{3}{l}{\textbf{Category 2: Reasoning \& Interpretive}} \\
\hline
Cause-Effect Reasoning & Infer what might have caused the current situation shown in the image. & “Explain what events might have led to the current scene.” \\
Temporal Reasoning & Predict what could happen before or after the moment captured in the image. & “Describe what is likely to happen next.” \\
Hypothesis Verification & Judge whether a given statement about the image is true or false. & “Is it true that the person is painting on a wall? Justify your answer.” \\
Explanation & Explain why certain visual elements appear or interact as they do. & “Why is the object on the left brighter than others?” \\
\hline
\multicolumn{3}{l}{\textbf{Category 3: Emotional \& Perspective}} \\
\hline
Mood Interpretation & Describe the emotional tone or atmosphere conveyed by the image. & “What kind of mood does this scene evoke, and why?” \\
Role-Play & Write from the viewpoint of a character or object in the image. & “Describe the scene from the perspective of the person sitting at the table.” \\
Empathy Reflection & Explain how the scene might make someone feel and why. & “How might a viewer feel when seeing this image, and what elements cause that feeling?” \\
\hline
\multicolumn{3}{l}{\textbf{Category 4: Creative \& Narrative}} \\
\hline
Story Continuation & Create a short story based on the visual context. & “Continue the story suggested by this scene in one paragraph.” \\
Imagination Extension & Imagine an alternative scenario that could occur in this setting. & “Describe how this scene would look if it took place at night.” \\
Artistic Captioning & Write a poetic or artistic caption inspired by the image. & “Compose a short, artistic caption that captures the feeling of the moment.” \\
Visual Poem & Compose a short poem describing the scene’s essence. & “Write a haiku inspired by the colors and atmosphere of this image.” \\
\hline
\multicolumn{3}{l}{\textbf{Category 5: Instructional \& Role-based}} \\
\hline
Visual Explanation & Explain a visual concept (e.g., geometry, physics, or design) using the image. & “Explain the concept of reflection as illustrated in this picture.” \\
Teaching Prompt & Use the image to instruct or teach a concept to someone. & “Use this image to teach a child what ‘balance’ means.” \\
Task Design & Design a new task or question related to the image content. & “Create an instruction-following question based on this image.” \\
\hline
\end{tabular}
\caption{
An overview of the 18 task templates defined in our VC-IFEngine, grouped into five major categories. 
Each template guides the model to perform a distinct type of instruction-following behavior grounded in visual context.
}
\label{tab:task_list}
\end{table*}

\begin{table*}[t]
\centering
\small
\setlength{\tabcolsep}{4pt}
\begin{tabular}{p{0.17\linewidth} p{0.30\linewidth} p{0.45\linewidth}}
\hline
\textbf{Constraint Type} & \textbf{Description} & \textbf{Example} \\
\hline
\multicolumn{3}{l}{\textbf{Category 1: Objective \& Analytical}} \\
\hline
Spatial & Emphasizes object positions, layouts, and relative arrangements within the scene. & “Describe only the objects located in the upper-right quadrant of the image.” \\
Attribute & Focuses on fine-grained visual properties such as color, texture, or material. & “Explain how the texture and material of the brick wall contribute to the scene.” \\
Comparative & Requires comparison between multiple objects or regions within the image. & “Compare the two vehicles in terms of size, color, and purpose.” \\
Counting \& Numerical & Involves precise object counting or quantitative reasoning. & “Count all the visible blue chairs and report the total.” \\
Textual (OCR) \& Grounding & Targets reading or interpreting textual information present in the image. & “Transcribe and explain the meaning of the text written on the billboard.” \\
\hline
\multicolumn{3}{l}{\textbf{Category 2: Inferential \& Creative}} \\
\hline
Causal \& Temporal & Focuses on reasoning about event order, causality, or temporal dynamics implied by the scene. & “Predict what is most likely to happen next after this moment.” \\
Affective \& Atmospheric & Highlights the emotional tone, mood, or atmosphere conveyed by the visual content. & “Analyze how lighting and color tone create a calm or tense mood.” \\
Perspective \& Role-Play & Alters the viewpoint or role from which the image should be interpreted. & “Describe the scene from the perspective of a child sitting on the swing.” \\
Hypothetical \& Counterfactual & Encourages imagining alternative or hypothetical scenarios. & “Explain how the scene’s atmosphere would change if it were nighttime.” \\
Abstract \& Conceptualization & Connects visual elements to abstract themes or ideas. & “Interpret how this image reflects the concept of ‘community’ or ‘isolation’.” \\
\hline
\end{tabular}
\caption{
Comprehensive overview of the constraint pool used in VC-IFEngine. 
}
\label{tab:constraint_pool}
\end{table*}

\clearpage
\begin{figure*}[p]
\centering
\begin{tcolorbox}[
    colback=blue!1!white,
    colframe=blue!70!black,
    title=Task Generation Prompt Templates,
    colbacktitle=blue!5!white,
    coltitle=blue!50!black,
    fonttitle=\bfseries,
    boxrule=0.5pt,
    arc=4pt,
    boxsep=5pt,
    left=6pt,
    right=6pt,
    top=6pt,
    bottom=6pt,
    breakable
]
\small
\raggedright

\textbf{(a) Screening Prompt: Task-Only Suitability}\\
You are given a task instruction $T$ and an image $I$. Decide whether $T$ is suitable for visual extension, meaning whether it is able to develop visual constraints that are relevant to $I$.\\
Answer “Yes” if $T$ is open-ended and general enough to be meaningfully extended with visual information or constraints in later stages.\\
Answer “No” if $T$ is overly specific, closed-form, or lacks potential for incorporating visual elements during extension.\\
\emph{Output:} Yes or No.\\[8pt]

\textbf{(b) Refinement Prompt (used when Yes)}\\
You are given the original task $T$ (screened as suitable) and an image $I$. Rewrite $T$ into a clearer and more specific form $T'$ that is tailored to $I$. The refined task should reference the image context at a high level (e.g., scene, salient objects, or setting) to improve clarity, but must not introduce new constraints. Keep $T'$ as a single concise sentence.\\
\emph{Output format:} Refined task: \textless one concise rewritten instruction grounded in $I$\textgreater.\\[8pt]

\textbf{(c) Replacement \& Refinement Prompt (used when No)}\\
You are given an image $I$ and an external task list $L=\{t_1,\dots,t_k\}$. Select one task $t^\ast \in L$ that is most relevant to $I$ and rewrite it into $T'$ so that it clearly fits the content of $I$. The refinement should tailor wording to the image context without adding any constraints. Keep $T'$ as a single concise sentence.\\
\emph{Output format:} Selected: \textless id or name of $t^\ast$\textgreater \quad Refined task: \textless one concise rewritten instruction grounded in $I$\textgreater.
\end{tcolorbox}

\caption{
An example of task generation prompt templates: (a) screen task suitability for visual extension; (b) if suitable, refine the same task using the given image; (c) if not, select an alternative from an external list and refine it using the image.
}
\label{fig:task_prompt_templates}
\end{figure*}
\clearpage
\clearpage
\begin{figure*}[t]
\centering
\begin{tcolorbox}[
    colback=blue!1!white,
    colframe=blue!70!black,
    title=Constraint Generation Prompt Templates,
    colbacktitle=blue!5!white,
    coltitle=blue!50!black,
    fonttitle=\bfseries,
    boxrule=0.5pt,
    arc=4pt,
    boxsep=5pt,
    left=6pt,
    right=6pt,
    top=6pt,
    bottom=6pt,
    breakable
]
\small
\raggedright

\textbf{(a) Constraint Selection Prompt}\\
You are given an image $I$, a task instruction $T$, and a predefined \emph{constraint pool} containing multiple types of visual reasoning constraints. Below listed the constraint pool: ...

Based on the semantics of $T$ and the visual content of $I$, select six constraint types that are most relevant and can enhance the visual grounding and diversity of the task.  
Each selected constraint should strengthen the connection between textual understanding and specific visual cues present in $I$.  
Output format: “Selected Constraints: [Type 1, Type 2, ..., Type 6]”.\\[8pt]

\textbf{(b) Constraint Refinement Prompt}\\
For each selected constraint type, refine it into an image-specific and semantically coherent form.  
The refined constraint should (1) clearly reference visual regions, entities, or relations within the image, and  
(2) remain consistent with the original task instruction.  
Example Output:
\begin{quote}
1. Focus on the arrangement of people and vehicles in the left half of the image.\\
2. Compare the density of the crowd in the foreground with that in the background.\\
3. Describe the scene from the viewpoint of a police officer near the barrier.\\
4. Pay attention to the color contrast between the street lights and the sky.\\
5. Count the number of vehicles visible on the main street.\\
6. Analyze how the lighting and atmosphere convey a sense of urban vibrancy.
\end{quote}
\vspace{4pt}

\textbf{(c) Quality Judgement Prompt}\\
Review all six refined constraints to ensure semantic coherence and logical compatibility.  
Check whether any constraints conflict, overlap excessively, or deviate from the main task focus.  
Output “Pass” if the constraint set is consistent; otherwise, describe detected issues and suggest minimal adjustments.
\end{tcolorbox}

\caption{
An example of the constraint generation prompt templates used in VC-IFEngine.  
Given the task, image, and predefined constraint pool, the model selects six relevant constraints, refines them into image-grounded formulations, and performs quality judgement to ensure semantic consistency and non-contradiction.
}
\label{fig:constraint_prompt_templates}
\end{figure*}
\clearpage

\begin{figure*}[t]
\centering
\begin{tcolorbox}[
    colback=blue!1!white,
    colframe=blue!70!black,
    title=VC-IFDPO Example (Rejected Response),
    colbacktitle=blue!5!white,
    coltitle=blue!50!black,
    fonttitle=\bfseries,
    boxrule=0.5pt,
    arc=4pt,
    boxsep=5pt,
    left=6pt,
    right=6pt,
    top=6pt,
    bottom=6pt,
    breakable
]
\small

\noindent
\vspace{8pt}

\raggedright
\small
\textbf{(d) Rejected}\\

This image shows a vibrant scene at a ski resort, with a busy atmosphere shaped by the presence of people, ski lifts, and snow covered terrain. The scene reflects an active winter environment where infrastructure and human activity are closely intertwined.

In the foreground, a group of individuals gathers near the base of a chairlift. This area appears relatively crowded, with skiers and possibly some spectators standing close together. Several people seem to be preparing their equipment, while others are waiting to board the chairlift. Ski poles and skis are visible, indicating that most individuals in this area are engaged in skiing related activities. The chairlift itself is prominently visible, with a chair suspended in the air, serving as a central focal point and a key transportation hub for moving skiers up the slope.

The midground displays a moderate density of people spread across the ski slopes. Individuals are more dispersed than in the foreground, suggesting that they are actively skiing, navigating the terrain, or taking short breaks. Some appear to be descending the slopes, while others move more cautiously. The cable system of the ski lifts extends into this region, visually connecting the foreground with the background. The chairs along the cable are spaced apart, indicating an organized and efficient flow rather than overcrowding.

In the background, the image opens into a wide expanse of snow covered slopes with people scattered across the terrain. The density of individuals decreases with distance from the chairlift area, emphasizing the openness of the environment. People in this region are likely skiing, snowboarding, or enjoying the slopes at a more relaxed pace. Their dispersed distribution suggests active movement combined with leisure.

The entire landscape is covered in snow, with gently rolling slopes extending into the distance. The terrain appears well maintained, with clear ski paths and designated areas visible across the scene. The uniform white surface of the snow highlights the contrast between the natural environment and human activity.

Signage and barriers are visible throughout the scene, particularly near the ski lift area and along parts of the slopes. Banners and flags likely mark boundaries, ski runs, or provide guidance for skiers. Colored barriers help separate different zones, such as areas for boarding the lift or spaces reserved for spectators.

Overall, the highest concentration of people is found near the ski lifts, especially in the foreground, where skiers gather before ascending or after descending. As people move farther away from these central points, their distribution becomes increasingly dispersed across the slopes. This pattern reflects the role of the ski lifts as key nodes of activity within the resort.

The overall atmosphere is lively and dynamic, combining moments of preparation, movement, and leisure. The snow covered landscape creates a calm visual backdrop, while the presence of ski lifts and active skiers adds a strong sense of motion and energy. Together, these elements form a well organized and functional skiing environment focused on winter recreation.
\end{tcolorbox}

\caption{
An example of VC-IFDPO data (rejected response)
}
\label{fig:dpo_rejected}
\end{figure*}

\begin{figure*}[t]
\centering
\begin{tcolorbox}[
    colback=blue!1!white,
    colframe=blue!70!black,
    title=Evaluation Prompt Templates for VC-IFEval,
    colbacktitle=blue!5!white,
    coltitle=blue!50!black,
    fonttitle=\bfseries,
    boxrule=0.5pt,
    arc=4pt,
    boxsep=5pt,
    left=6pt,
    right=6pt,
    top=6pt,
    bottom=6pt,
    breakable
]
\small
\raggedright

\textbf{(a) Direct Judge Prompt.}\\
You are asked to judge whether the AI assistant’s response fully complies with each listed constraint. \
Follow the evaluation principles below carefully:

1. Apply a consistent and rigorous standard when making your decisions.  
2. Each judgment should be grounded in the visual evidence provided by the image.  
3. For every constraint, assign 1 point if it is completely satisfied; assign 0 otherwise.

<start of response>
{prediction}
<end of response>

<start of constraint list>
constraints
<end of constraint list>

Evaluate every constraint separately and provide a short explanation for each decision. \
Do not skip or merge any constraints. After completing all evaluations, give an overall summary that lists the scores for every constraint in one concise line.

Your output format must be exactly as follows:
Judgement: ...
Summary: constraint\_1: x/1, constraint\_2: x/1, constraint\_3: x/1, ..., constraint\_n: x/1.

\textbf{(b) Comparative Judge Prompt.}\\
You are evaluating whether the availability of IMAGE caused a substantive influence on the model’s answer.
You will be given the question and two answers:
- Answer A: produced WITH image available.
- Answer B: produced WITHOUT image.

Guidelines:
- If Answer A contains details that plausibly come from visual evidence (objects, layout, colors, counts, attributes) and such details are missing/incorrect in Answer B, or the final conclusions differ BECAUSE of visual cues, judge it as "Influenced".
- If both answers are essentially the same in conclusions and key details (only minor wording differs), judge "Not influenced".
- Base your judgment strictly on the textual differences and the question. Do NOT assume seeing the image yourself.

Question: {question}

Answer A (WITH image): 

Answer B (WITHOUT image):

Return exactly one word: Influenced or Not influenced.
\end{tcolorbox}

\caption{
Prompt templates for the evaluation stage in VC-IFEval.  
(a) \textbf{Direct Judge} assesses a single response for visual and instruction-following fidelity.  
(b) \textbf{Comparative Judge} compares responses generated with and without visual input to isolate the effect of linguistic priors and assess whether the model’s improvement stems from genuine visual grounding.
}
\label{fig:vc_ifeval_judge_templates}
\end{figure*}

\begin{figure*}[t]
\centering
\begin{tcolorbox}[
    colback=blue!1!white,
    colframe=blue!70!black,
    title=Human Inter-Annotator Agreement Example,
    colbacktitle=blue!5!white,
    coltitle=blue!50!black,
    fonttitle=\bfseries,
    boxrule=0.5pt,
    arc=4pt,
    boxsep=5pt,
    left=6pt,
    right=6pt,
    top=6pt,
    bottom=6pt,
    breakable
]
\small

\noindent
\begin{minipage}[t]{0.44\linewidth}
    \centering
    \vspace{0pt}
    \includegraphics[width=\linewidth]{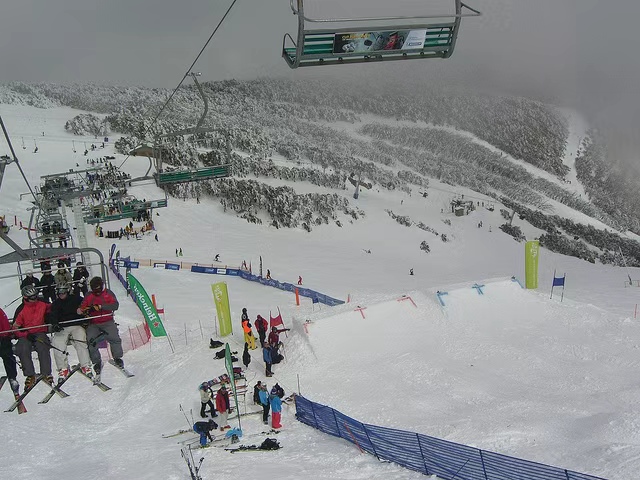}
    \vspace{4pt}
    {\small \textit{(a) Image.}}
\end{minipage}
\hfill
\begin{minipage}[t]{0.53\linewidth}
    \vspace{0pt}
    \raggedright
    \small
    {\bf (b) Task \& Constraints}\\[4pt]
    \textbf{Task.} Describe the picture, focusing on the spatial arrangement of
people, ski lifts, and snow-covered terrain.\\[3pt]
    \textbf{Constraints.}
    \begin{itemize}\itemsep 1pt
        \item Analyze how individuals in the foreground, are engaged in activities.
        \item Compare the density of people near the ski lifts.
        \item Predict how the flow of movement and activity might change if the ski lift were to stop.
        \item Consider the overall mood and atmosphere.
        \item Imagine how the scene might evolve if the event were to shift to a different time of day.
    \end{itemize}
\end{minipage}

\vspace{10pt}
\noindent\begin{minipage}[t]{0.95\linewidth}
    \centering
    \small
    {\bf (c) Model Response}\\[3pt]
    \begin{minipage}{0.88\linewidth}
        \small
        ...
        In the foreground, several individuals near the base of a chairlift appear to be preparing their equipment, adjusting ski gear, or waiting to board the chairlift.
...
The presence of ski poles and ski boards suggests they are skiers engaged in active preparation.
...
The density of people is highest near the ski lifts, particularly in the foreground.
This area acts as a convergence point where skiers gather before ascending the slope or after descending.
...
As people move away from the lifts, their distribution becomes more dispersed.
...
The atmosphere is dynamic and lively, with a mix of activity and leisure.
The snow-covered landscape creates a serene yet energetic environment.
...
The presence of the ski lifts and scattered individuals adds a sense of movement and excitement typical of a busy ski resort.
...
    \end{minipage}
\end{minipage}

\vspace{10pt}
\noindent
\begin{minipage}[t]{0.47\linewidth}
    \raggedright
    \small
    {\bf (d) Human A}\\[4pt]
    \begin{itemize}\itemsep 1pt
        \item \textbf{Constraint Coverage:} 1 (1), 2 (1), 3 (0), 4 (1), 5(0)
    \end{itemize}
\end{minipage}
\hfill
\begin{minipage}[t]{0.47\linewidth}
    \raggedright
    \small
    {\bf (e) Human B}\\[4pt]
    \begin{itemize}\itemsep 1pt
        \item \textbf{Constraint Coverage:} 1 (1), 2 (1), 3 (0), 4 (0), 5(0)
    \end{itemize}
\end{minipage}

\vspace{6pt}
\noindent\small
Agreement summary: Cohen's $\kappa=0.62$ (substantial agreement), inter-annotator agreement = 0.80 on constraint satisfaction labels

\end{tcolorbox}

\caption{Human inter-annotator agreement layout for a single image--task instance.}
\label{fig:agreement}
\end{figure*}

\end{document}